\documentclass[10pt,twocolumn,letterpaper]{article}

\usepackage{cvpr}
\usepackage{times}
\usepackage{epsfig}
\usepackage{graphicx}
\usepackage{amsmath}
\usepackage{amssymb}
\newcommand{\tabincell}[2]{\begin{tabular}{@{}#1@{}}#2\end{tabular}}

\cvprfinalcopy % *** Uncomment this line for the final submission

\def\cvprPaperID{****} % *** Enter the CVPR Paper ID here

\begin{document}

%%%%%%%%% TITLE
\title{Learning for Video Super-Resolution through HR Optical Flow Estimation}

\author{Longguang Wang, Yulan Guo, Zaiping Lin, Xinpu Deng, and Wei An\\
School of Electronic Science, National University of Defense Technology\\
Changsha 410073, China\\
{\tt\small \{wanglongguang15, yulan.guo, linzaiping, dengxinpu, anwei\}@nudt.edu.cn}
}

\maketitle
%\thispagestyle{empty}

%%%%%%%%% ABSTRACT
\begin{abstract}
	Video super-resolution (SR) aims to generate a sequence of high-resolution (HR) frames with plausible and temporally consistent details from their low-resolution (LR) counterparts. The generation of accurate correspondence plays a significant role in video SR. It is demonstrated by traditional video SR methods that simultaneous SR of both images and optical flows can provide accurate correspondences and better SR results. However, LR optical flows are used in existing deep learning based methods for correspondence generation. In this paper, we propose an end-to-end trainable video SR framework to super-resolve both images and optical flows. Specifically, we first propose an optical flow reconstruction network (OFRnet) to infer HR optical flows in a coarse-to-fine manner. Then, motion compensation is performed according to the HR optical flows. Finally, compensated LR inputs are fed to a super-resolution network (SRnet) to generate the SR results. Extensive experiments demonstrate that HR optical flows provide more accurate correspondences than their LR counterparts and improve both accuracy and consistency performance. Comparative results on the Vid4 and DAVIS-10 datasets show that our framework achieves the state-of-the-art performance. The codes will be released soon at: https://github.com/LongguangWang/SOF-VSR-Super-Resolving-Optical-Flow-for-Video-Super-Resolution-. 
\end{abstract}

%%%%%%%%% BODY TEXT
\section{Introduction}
Super-resolution (SR) aims to generate high-resolution (HR) images
or videos from their low-resolution (LR) counterparts. As a typical
low-level computer vision problem, SR has been widely investigated for decades \cite{2001-AComputationallyEfficientSuperresolutionImageReconstructionAlgorithm-Nguyen-573-583,2007-ImageUpsamplingViaImposedEdgeStatistics-Fattal-95-95,2011-ImageandVideoUpscalingfromLocalSelfExamples-Freedman-1-12}.
Recently, the prevalence of high-definition display further advances
the development of SR. For single image SR, image details are recovered using the spatial correlation in a single frame. In contrast, inter-frame temporal
correlation can further be exploited for video SR.

Since temporal correlation is crucial to video SR, the key to success lies in accurate correspondence generation. Numerous methods \cite{2007-OpticalFlowBasedSuperResolution:aProbabilisticApproach-Fransens-106-115,2014-OnBayesianAdaptiveVideoSuperResolution-Liu-346-360,2015-HandlingMotionBlurinMultiFrameSuperResolution-Ma-5224-5232} have demonstrated that the correspondence generation and SR problems are closely interrelated and can boost each other's accuracy. Therefore, these methods integrate the SR of both images and optical flows in a unified framework. However, current deep learning based methods \cite{2015-VideoSuperResolutionViaDeepDraftEnsembleLearning-Liao-531-539,2016-VideoSuperResolutionwithConvolutionalNeuralNetworks-Kappeler-109-122,2017-DetailRevealingDeepVideoSuperResolution-Tao-4482-4490,2017-RealTimeVideoSuperResolutionwithSpatioTemporalNetworksandMotionCompensation-Caballero-2848-2857,2017-RobustVideoSuperResolutionwithLearnedTemporalDynamics-Liu--,2018-LearningTemporalDynamicsforVideoSuperResolution:aDeepLearningApproach-Liu--} mainly focus on the SR of images, and use LR optical flows to provide correspondences. Although LR optical flows can provide sub-pixel correspondences in LR images, their limited accuracy hinders the performance improvement for video SR, especially for scenarios with large upscaling factors.

\begin{figure}[bt]
\centering
\includegraphics[width=0.8\linewidth]{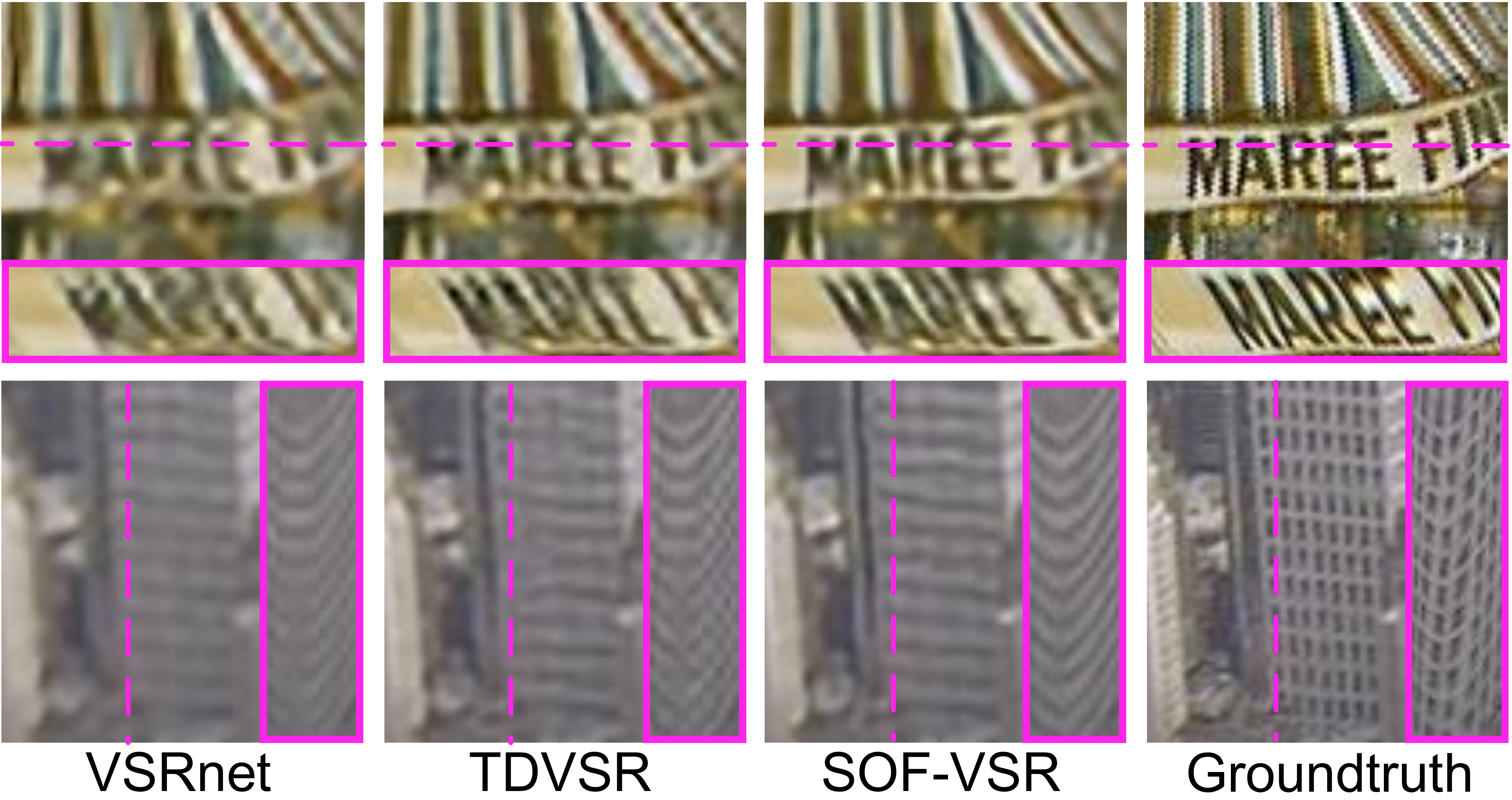}
\caption{Temporal profiles under $\times4$ configuration for VSRnet \cite{2016-VideoSuperResolutionwithConvolutionalNeuralNetworks-Kappeler-109-122}, TDVSR \cite{2017-RobustVideoSuperResolutionwithLearnedTemporalDynamics-Liu--} and our SOF-VSR on \emph{Calendar} and \emph{City}. Purple boxes represent corresponding temporal profiles. Our SOF-VSR produces finer details in temporal profiles, which are more consistent with the groundtruth.}
\label{fig1}
\end{figure}

\begin{figure*}[tb]
\centering
\includegraphics[width=0.8\linewidth]{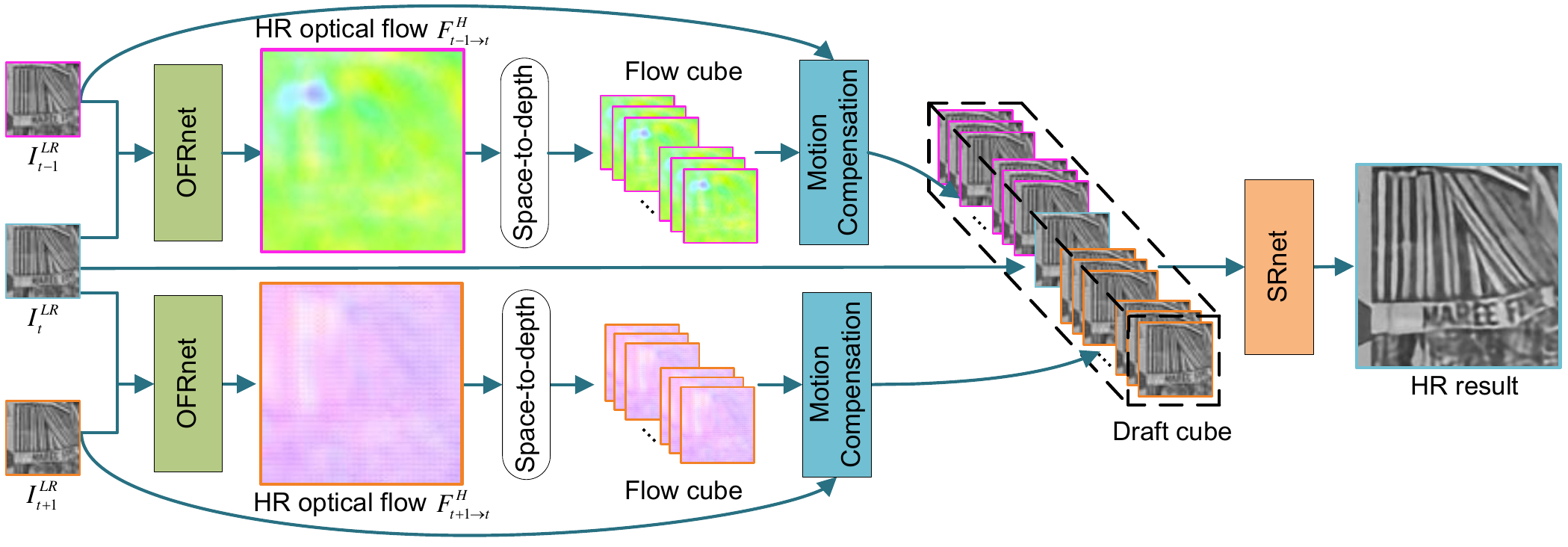}
\caption{Overview of the proposed framework. Our framework is fully convolutional and can be trained in an end-to-end manner.}
\label{fig2}
\end{figure*}

In this paper, we propose an end-to-end trainable video SR framework to generate both HR images and optical flows. The SR of optical flows provides accurate correspondences, which not only improves the accuracy of each HR image, but also achieves better temporal consistency. We first introduce an optical flow reconstruction net (OFRnet) to reconstruct HR optical flows in a coarse-to-fine manner. These HR optical flows are then used to perform motion compensation on LR frames. A space-to-depth transformation is therefore used to bridge the resolution gap between HR optical flows and LR frames. Finally, the compensated LR frames are fed to a super-resolution net (SRnet) to generate each HR frame. Extensive evaluation is conducted to test our framework. Comparison to existing video SR methods shows that our framework achieves the state-of-the-art performance in terms of peak signal-to-noise ratio (PSNR) and structural similarity index (SSIM). Moreover, our framework achieves better temporal consistency for visual
perception (as shown in Fig.~\ref{fig1}).

Our main contributions can be summarized as follows: 1) We integrate the SR of both images and optical flows into a single SOF-VSR (super-resolving optical flow for video SR) network. The SR of optical flows provides accurate correspondences and improves the overall performance; 2) We propose an OFRnet to infer HR optical flows in a coarse-to-fine manner; 3) Extensive experiments have demonstrated the effectiveness of our framework. It is shown that our framework achieves the state-of-the-art performance.

\section{Related Work}
In this section, we briefly review some major methods for single image SR and video SR.

\subsection{Single Image SR}
Dong \emph{et al.} \cite{2014-LearningaDeepConvolutionalNetworkforImageSuperResolution-Dong-184-199} proposed the pioneering work to use deep learning for single image SR. They used a three-layer convolutional neural network (CNN) to approximate the non-linear mapping from the LR image to the HR image. Recently, deeper and more complex network architectures have been proposed \cite{2016-AccurateImageSuperResolutionUsingVeryDeepConvolutionalNetworks-Kim-1646-1654,2017-ImageSuperResolutionViaDeepRecursiveResidualNetwork-Tai-2790-2798,2018-FastandAccurateSingleImageSuperResolutionViaInformationDistillationNetwork-Hui--}. Kim \emph{et al.}\cite{2016-AccurateImageSuperResolutionUsingVeryDeepConvolutionalNetworks-Kim-1646-1654} proposed a very deep super-resolution network (VDSR) with 20 convolutional layers. Tai \emph{et al.} \cite{2017-ImageSuperResolutionViaDeepRecursiveResidualNetwork-Tai-2790-2798} developed a deep recursive residual network (DRRN) and used recursive learning to control the model parameters while increasing the depth. Hui \emph{et al.} \cite{2018-FastandAccurateSingleImageSuperResolutionViaInformationDistillationNetwork-Hui--} proposed an information distillation network to reduce computational complexity and memory consumption.

\begin{figure*}[htb]
	\centering
	\includegraphics[width=0.8\linewidth]{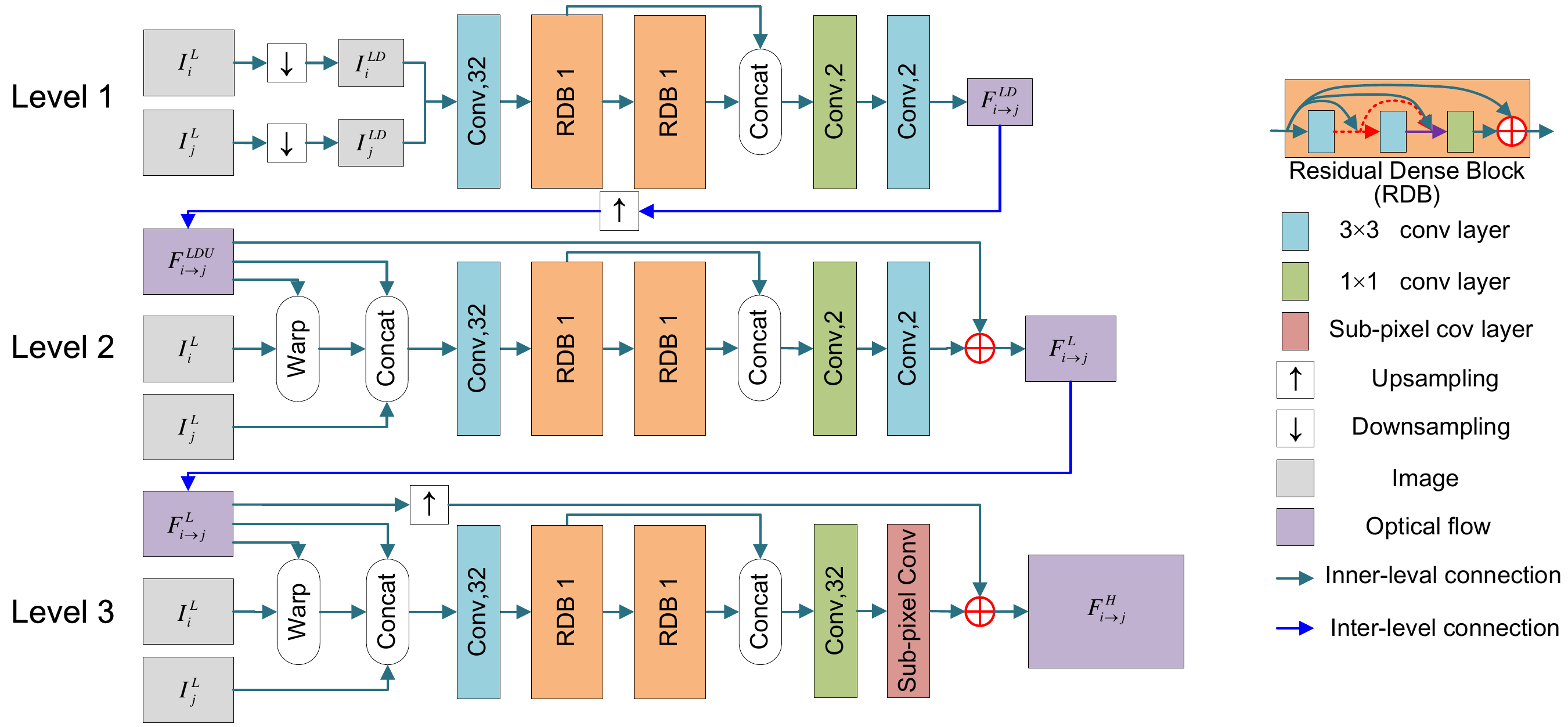}
	\caption{Architecture of our OFRnet. Our OFRnet works in a coarse-to-fine manner. At each level, the output of its previous level is used 
		to compute a residual optical flow.}
	\label{fig3}
\end{figure*}

\subsection{Video SR}
\noindent\textbf{Traditional video SR.} To handle complex motion patterns in video sequences, Protter \emph{et al.} \cite{2008-GeneralizingtheNonlocalMeanstoSuperResolutionReconstruction-Protter-36-51} generalized the non-local means framework to address the SR problem. They used patch-wise spatio-temporal similarity to perform adaptive fusion of multiple frames. Takeda \emph{et al.} \cite{2009-SuperResolutionwithoutExplicitSubpixelMotionEstimation-Takeda-1958-1975} further
introduced 3D kernel regression to exploit patch-wise spatio-temporal neighboring
relationship. However, the resulting HR images of these two methods are over-smoothed. To exploit pixel-wise correspondences, optical flow is used in \cite{2007-OpticalFlowBasedSuperResolution:aProbabilisticApproach-Fransens-106-115,2014-OnBayesianAdaptiveVideoSuperResolution-Liu-346-360,2015-HandlingMotionBlurinMultiFrameSuperResolution-Ma-5224-5232}. Since the accuracy of correspondences provided by optical flows in LR images is usually low \cite{2013-SimultaneousSuperResolutionofDepthandImagesUsingaSingleCamera-Lee-281-288}, an iterative framework is used in these methods \cite{2007-OpticalFlowBasedSuperResolution:aProbabilisticApproach-Fransens-106-115,2014-OnBayesianAdaptiveVideoSuperResolution-Liu-346-360,2015-HandlingMotionBlurinMultiFrameSuperResolution-Ma-5224-5232} to estimate both HR images and optical flows.

\noindent\textbf{Deep video SR with separated motion compensation.} Recently, deep learning has been investigated for video SR. Liao \emph{et al.} \cite{2015-VideoSuperResolutionViaDeepDraftEnsembleLearning-Liao-531-539}
performed motion compensation under different parameter settings to
generate an ensemble of SR-drafts, and then employed a CNN to recover
high-frequency details from the ensemble. Kappelar \emph{et al.} \cite{2016-VideoSuperResolutionwithConvolutionalNeuralNetworks-Kappeler-109-122}
also performed image alignment through optical flow estimation, and
then passed the concatenation of compensated LR inputs to a CNN to reconstruct
each HR frame. In these methods, motion compensation is separated from CNN. Therefore, it is difficult for them to obtain the overall optimal solution.

\noindent\textbf{Deep video SR with integrated motion compensation.} More recently, Caballero \emph{et al.} \cite{2017-RealTimeVideoSuperResolutionwithSpatioTemporalNetworksandMotionCompensation-Caballero-2848-2857}
proposed the first end-to-end CNN framework (namely, VESPCN) for video SR. It comprises
a motion compensation module and a sub-pixel
convolutional layer used in \cite{2016-RealTimeSingleImageandVideoSuperResolutionUsinganEfficientSubPixelConvolutionalNeuralNetwork-Shi-1874-1883}. Since that, end-to-end framework with motion compensation
dominates the research of video SR. Tao \emph{et al.} \cite{2017-DetailRevealingDeepVideoSuperResolution-Tao-4482-4490}
used the motion estimation module in VESPCN, and proposed an encode-decoder network based on LSTM. This architecture facilitates the extraction of temporal context. Liu \emph{et al.} \cite{2017-RobustVideoSuperResolutionwithLearnedTemporalDynamics-Liu--} customized ESPCN \cite{2016-RealTimeSingleImageandVideoSuperResolutionUsinganEfficientSubPixelConvolutionalNeuralNetwork-Shi-1874-1883} to
simultaneously process different numbers of LR frames. They then introduced a temporal adaptive network to aggregate multiple HR
estimates with learned dynamic weights. Sajjadi \emph{et al.} \cite{2018-FrameRecurrentVideoSuperResolution-Sajjadi--} proposed a frame-recurrent architecture to use previously inferred HR estimates for the SR of subsequent frames. The recurrent architecture can assimilate previous inferred HR frames without increase in computational demands.

It is already demonstrated by traditional video SR methods \cite{2007-OpticalFlowBasedSuperResolution:aProbabilisticApproach-Fransens-106-115,2014-OnBayesianAdaptiveVideoSuperResolution-Liu-346-360,2015-HandlingMotionBlurinMultiFrameSuperResolution-Ma-5224-5232} that simultaneous SR of images and optical flows produces better result. However, current CNN-based methods only focus on the SR of images. Different from previous works, we propose an end-to-end video SR framework to super-resolve both images and optical flows. It is demonstrated that the SR of optical flows facilitates our framework to achieve the state-of-the-art performance.

\section{Network Architecture}

Our framework takes $N$ consecutive LR frames as inputs and super-resolves
the central frame. The LR inputs are first divided into pairs and fed
to OFRnet to infer an HR optical flow. Then, a space-to-depth
transformation \cite{2018-FrameRecurrentVideoSuperResolution-Sajjadi--}
is employed to shuffle the HR optical flow into LR grids. Afterwards,
motion compensation is performed to generate an LR draft cube. Finally, the draft cube is fed to SRnet to infer the HR frame. The overview of
our framework is shown in Fig.~\ref{fig2}.

\subsection{Optical Flow Reconstruction Net (OFRnet)}

It is demonstrated that CNN has the capability to learn the non-linear mapping between LR and HR images for the SR problem \cite{2014-LearningaDeepConvolutionalNetworkforImageSuperResolution-Dong-184-199}. Recent CNN-based works \cite{2015-FlowNet:LearningOpticalFlowwithConvolutionalNetworks-Dosovitskiy--,2017-FlowNet2.0:EvolutionofOpticalFlowEstimationwithDeepNetworks-Ilg--} have also shown the potential for motion estimation. In this paper, we incorporate these two tasks into a unified network to infer HR optical flows from LR images. Specifically, our OFRnet takes a pair of LR frames $I_{i}^{L}$ and $I_{j}^{L}$
as inputs, and reconstruct an optical flow between their corresponding
HR frames $I_{i}^{H}$ and $I_{j}^{H}$:

\begin{equation}
F_{i\rightarrow j}^{H}=\mathbf{Net}_{OFR}(I_{i}^{L},\,I_{j}^{L};\,\Theta_{OFR})
\end{equation}
where $F_{i\rightarrow j}^{H}$ represents the HR optical flow and $\Theta_{OFR}$ is
the set of parameters.

Motivated by the pyramid optical flow estimation method in \cite{1999-PyramidalImplementationoftheLucasKanadeFeatureTracker:DescriptionoftheAlgorithm-Bouguet--},
we use a coarse-to-fine approach to handle complex motion patterns (especially large displacements). As illustrated
in Fig.~\ref{fig3}, a 3-level pyramid is employed in our OFRnet.

\textbf{Level 1:} The pair of LR images $I_{i}^{L}$
and $I_{j}^{L}$ are downsampled by a factor of 2 to produce
$I_{i}^{LD}$ and $I_{j}^{LD}$, which are further concatenated and fed to
a feature extraction layer. Then, two residual dense blocks (RDB) \cite{2018-ResidualDenseNetworkforImageSuperResolution-Zhang--}
with 4 layers and a growth rate of 32 are customized. Within each residual dense block, the first 3 layers
are followed by a leaky ReLU using a leakage factor of 0.1, while the last layer performs feature fusion. The residual dense block works in a local residual learning manner with a local skip connection at the end. Once dense features are extracted
by the residual dense blocks, they are concatenated and fed to a feature fusion layer. Then, the optical flow $F_{i\rightarrow j}^{LD}$ at this level is inferred by the subsequent flow estimation layer.

\textbf{Level 2:} Once the raw optical flow $F_{i\rightarrow j}^{LD}$
is obtained from level 1, it is upscaled by a factor
of 2. The upscaled flow $F_{i\rightarrow j}^{LDU}$ is then used to warp $I_{i}^{L}$, resulting in $\hat{I}_{i\rightarrow j}^{L}$. Next, $\hat{I}_{i\rightarrow j}^{L}$, $I_{j}^{L}$ and $F_{i\rightarrow j}^{LDU}$ are concatenated and fed
to a network module. Note that, this module at level 2 is similar to that at level 1, except that residual learning is used.

\textbf{Level 3:} The module at level 2 generates an optical flow $F_{i\rightarrow j}^{L}$ with the same size as the LR input $I_{j}^{L}$. Therefore, the module at level 3 works as an SR part to infer the HR optical flow. The architecture at level 3 is similar to level 2 except that
the flow estimation layer is replaced by a sub-pixel convolutional layer \cite{2016-RealTimeSingleImageandVideoSuperResolutionUsinganEfficientSubPixelConvolutionalNeuralNetwork-Shi-1874-1883}
for resolution enhancement.

\begin{figure}[htb]
	\centering
	\includegraphics[width=0.8\linewidth]{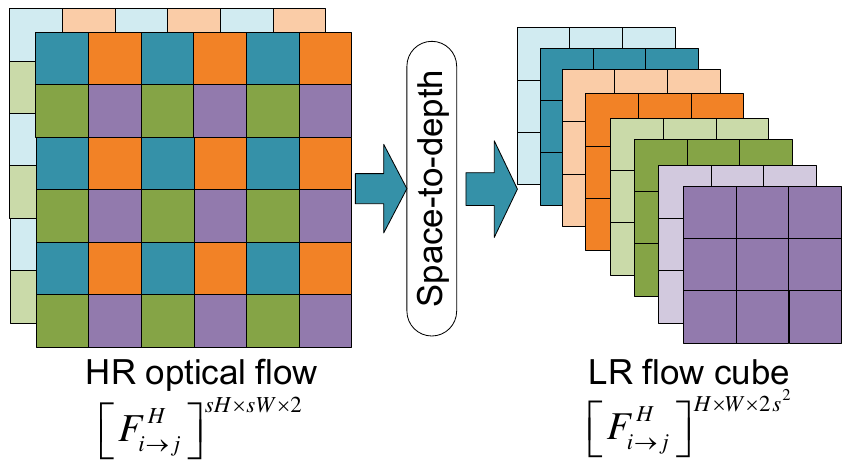}
	\caption{Illustration of space-to-depth transformation. The space-to-depth transformation folds an HR optical flow in LR space to generate an LR flow cube.}
	\label{fig4}
\end{figure}

Although numerous networks for SR \cite{2017-EnhanceNet:SingleImageSuperResolutionthroughAutomatedTextureSynthesis-Sajjadi-4501-4510,2017-DeepLaplacianPyramidNetworksforFastandAccurateSuperResolution-Lai-5835-5843,2017-ImageSuperResolutionViaDeepRecursiveResidualNetwork-Tai-2790-2798,2018-FastandAccurateSingleImageSuperResolutionViaInformationDistillationNetwork-Hui--} and optical flow estimation \cite{2017-PWCNet:CNNsforOpticalFlowUsingPyramidWarpingandCostVolume-Sun--,2017-OpticalFlowEstimationUsingaSpatialPyramidNetwork-Ranjan-2720-2729,2018-LiteFlowNet:aLightweightConvolutionalNeuralNetworkforOpticalFlowEstimation-Hui--} can be found in literature, our OFRnet is, to the best of our knowledge, the first unified network to integrate these two tasks. Note that, inferring HR optical flow from LR images is quite challenging, our OFRnet has demonstrated the potential of CNN to address this challenge. Our OFRnet is compact, with only 0.6M parameters. It is further demonstrated in Sec.~\ref{section4_3} that the resulting HR optical flows benefit our video SR framework in both accuracy and consistency performance.

\subsection{Motion Compensation}

Once HR optical flows are produced by OFRnet, space-to-depth transformation is used to bridge the resolution gap between HR optical flows and
LR frames. As illustrated in Fig.~\ref{fig4}, regular LR grids are extracted
from the HR flow and placed into the channel dimension to derive
a flow cube with the same resolution as LR frames:
\begin{equation}
\left[F_{i\rightarrow j}^{H}\right]^{sH\times sW\times2}\rightarrow\left[F_{i\rightarrow j}^{H}\right]^{H\times W\times2s^{2}}
\end{equation}
where $H$ and $W$ represent the size of the LR frame, $s$ is the upscaling factor. Note that, the magnitude of optical flow is divided by a scalar $s$ during the transformation to match the spatial resolution of LR frames.

Then, slices are extracted from the flow
cube to warp the LR frame $I_{i}^{LR}$, resulting in multiple warped drafts:
\begin{equation}
C_{i\rightarrow j}^{L}=\textup{W}(I_{i}^{L},\,\left[F_{i\rightarrow j}^{H}\right]^{H\times W\times2s^{2}})
\end{equation}
where $\textup{W}(\cdot)$ denotes warping operation and $C_{i\rightarrow j}^{L}\!\in\!R^{H\times W\times{s^{2}}}$ represents
the warped drafts after concatenation, namely draft cube. 

\subsection{Super-resolution Net (SRnet)}

After motion compensation, all the drafts are concatenated with the central LR frame, as shown in Fig.~\ref{fig2}. Then, the draft cube is fed to SRnet to infer the HR frame:
\begin{equation}
I_0^{SR}=\mathbf{Net}_{SR}(C^{L};\,\Theta_{SR})
\end{equation}
where $I_{0}^{SR}$ is the super-resolved result of the central
LR frame, $C^{L}\!\in\!R^{H\times W\times(2s^{2}+1)}$ represents the draft cube and $\Theta_{SR}$
is the set of parameters.

\begin{figure}[bt]
	\centering
	\includegraphics[width=0.9\linewidth]{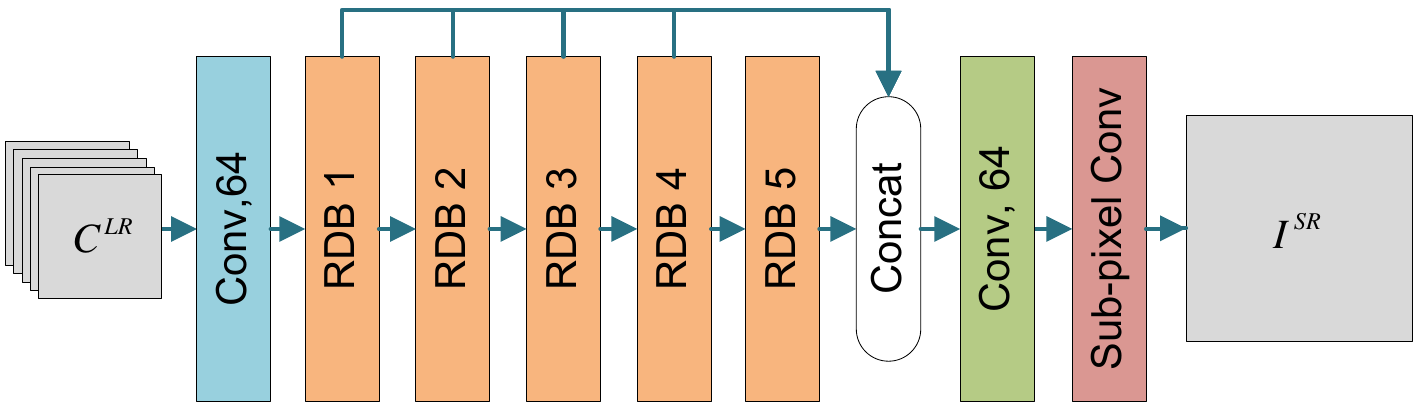}
	\caption{Architecture of our SRnet.}
	\label{fig5}
\end{figure}

As illustrated in Fig.~\ref{fig5}, the
draft cube is first passed to a feature extraction layer with 64 kernels, and then the output features are fed to 5 residual dense blocks (which are similar to our OFRnet).
Here, we increase the number of layers to 5 and the growth rate to 64 for each residual dense block. Afterwards, we concatenate all the outputs of  residual dense blocks and use a feature fusion layer to distillate the dense features.
Finally, a sub-pixel layer is used
to generate the HR frame.

The combination of densely connected layers and residual learning
in residual dense blocks has been demonstrated to have a contiguous memory mechanism
\cite{2018-ResidualDenseNetworkforImageSuperResolution-Zhang--,2017-DenselyConnectedConvolutionalNetworks-Huang-2261-2269}.
Therefore, we employ residual dense blocks in our SRnet to facilitate effective feature learning
from preceding and current local features. Furthermore, feature
reuse in the residual dense blocks improves the model compactness
and stabilizes the training process.

\subsection{Loss Function}

We design two loss terms $\mathcal{L_{\mathrm{OFR}}}$ and
$\mathcal{L_{\mathrm{SR}}}$ for OFRnet and SRnet, respectively. For the
training of OFRnet, intermediate supervision is used at each level
of the pyramid:
\begin{equation}
\mathcal{L_{\mathrm{OFR}}}\!=\!\sum_{i\in[-T,\,T],\,i\neq0}\!\frac{
	\mathcal{L}_{{level 3},i}\!+\!\lambda_{2}\mathcal{L}_{{level 2},i}+\!\lambda_{1}
	\mathcal{L}_{{level 1},i}}{2T}
\end{equation}
where
\begin{equation}
\left\{
\begin{aligned}
\mathcal{L}_{{level 3},i}&\!=\! \left\Vert \textup{W}(I_{i}^{H},\,F_{i\rightarrow0}^{H})\!-\!I_{0}^{H}
\right\Vert _{2}^{2}\!+\!\lambda_{3} \left\Vert \nabla F_{i\rightarrow0}^{H}\right\Vert _{1} \\
\mathcal{L}_{{level 2},i}&\!=\! \left\Vert \textup{W}(I_{i}^{L},\,F_{i\rightarrow0}^{L})\!-\!I_{0}^{L}
\right\Vert _{2}^{2}\!+\!\lambda_{3} \left\Vert \nabla F_{i\rightarrow0}^{L}\right\Vert _{1} \\
\mathcal{L}_{{level 1},i}&\!=\!\left\Vert \textup{W}(I_{i}^{LD}\!,\,F_{i\rightarrow0}^{LD})\!-\!I_{0}^{LD}\!\!
\right\Vert _{2}^{2} \!+\!\lambda_{3} \left\Vert \nabla F_{i\rightarrow0}^{LD}\right\Vert _{1}
\end{aligned}
\right.
\label{equ5}
\end{equation}
here $T$ denotes the temporal
window size and $\left\Vert \nabla F_{i\rightarrow0}^{H}\right\Vert _{1}$ is the regularization term to constrain the smoothness of the optical flow.
We empirically set $\lambda_{2}=0.25$ and $\lambda_{1}=0.125$ to make our OFRnet
focus on the last level. We also set $\lambda_{3}=0.01$ as the regularization coefficient.

For the training of SRnet, we use the widely applied mean square error (MSE) loss:

\begin{equation}
\mathcal{L_{\mathrm{SR}}}=\left\Vert I_0^{SR}-I_0^{H}\right\Vert _{2}^{2}
\end{equation}

Finally, the total loss used for joint training is $\mathcal{\mathcal{L}=L_{\mathrm{SR}}}+\lambda_{4}\mathcal{L_{\mathrm{OFR}}}$,
where $\lambda_{4}$ is empirically set to 0.01 to balance the two loss terms.

\section{Experiments}
In this section, we first conduct ablation experiments to evaluate our framework. Then, we further compare our framework to several existing video SR methods.

\subsection{Datasets}

We collected 152 1080P HD video clips from the CDVL Database\footnote{www.cdvl.org}
and the Ultra Video Group Database\footnote{ultravideo.cs.tut.fi}. The collected videos cover diverse natural and urban scenes. We used 145 videos from the CDVL Database as the training set, and 7 videos from the Ultra Video Group Database as the validation set.
Following the configuration in \cite{2014-OnBayesianAdaptiveVideoSuperResolution-Liu-346-360,2015-VideoSuperResolutionViaDeepDraftEnsembleLearning-Liao-531-539,2017-DetailRevealingDeepVideoSuperResolution-Tao-4482-4490},
we downsampled the video clips to the size of $540\times960$ as the HR groundtruth using Matlab $imresize$ function. In this paper, we only focus on the upscaling factor of 4 since it is the most challenging case. Therefore, the HR video clips were further downsampled to produce LR inputs of size $135\times240$. 

For fair comparison to the state-of-the-arts, we chose the widely
used Vid4 benchmark dataset. We also used another 10 video clips from the DAVIS dataset \cite{2017-The2017DAVISChallengeonVideoObjectSegmentation-Pont-Tuset--} for further comparison, which we refer to as DAVIS-10.

\begin{table*}[bt]
	\caption{Comparative results achieved by our framework and its variants on the Vid4 dataset under $\times4$	configuration. Best results are shown in boldface.}
	\label{tab1}
	\begin{center}
		\scriptsize
		\setlength{\tabcolsep}{0.3mm}{
			\begin{tabular}{lccccc}
				\hline 
				& PSNR($\uparrow$)  & SSIM($\uparrow$) & \tabincell{c}{T-MOVIE($\downarrow$)\\($\times10^{-3}$)} & \tabincell{c}{MOVIE($\downarrow$) \\($\times10^{-3}$)}  & VQM-VFD($\downarrow$)  \tabularnewline
				\hline
				SOF-VSR w/o OFRnet  & 25.80  & 0.760  & 20.08 & 4.54 & 0.240\tabularnewline
				SOF-VSR w/o OFRnet$\mathcal{_\mathrm{\emph{level3}}}$ & 25.88 & 0.764 & 19.95 & 4.48 & 0.235 \tabularnewline
				SOF-VSR w/o OFRnet$\mathcal{_\mathrm{\emph{level3}}}$ + upsampling & 25.86 & 0.763 & 19.92 & 4.50 & 0.231 \tabularnewline
				SOF-VSR & \textbf{26.01} & \textbf{0.771} & \textbf{19.78} & \textbf{4.32} & \textbf{0.227}\tabularnewline
				\hline
		\end{tabular}}
	\end{center}
\end{table*}

\subsection{Implementation Details}

Following  \cite{2014-LearningaDeepConvolutionalNetworkforImageSuperResolution-Dong-184-199,2017-RobustVideoSuperResolutionwithLearnedTemporalDynamics-Liu--}, 
we converted input LR frames into YCbCR color space and only fed the luminance channel to our network. All metrics in this section are computed in the luminance channel. During the training phase,
we randomly extracted 3 consecutive frames from an LR video clip, and randomly cropped a $32\times32$ patch as the input. Meanwhile, its corresponding
patch in HR video clip was cropped as groundtruth. Data augmentation
was performed through rotation and reflection to improve the generalization
ability of our network. 

We implemented our framework in PyTorch. We applied the Adam solver \cite{2015-Adam:aMethodforStochasticOptimization-Kingma--}
with $\beta_{1}=0.9$, $\beta_{2}=0.999$ and batch size of 16. The
initial learning rate was set to $10^{-4}$ and reduced to half after every 50K iterations.
We trained our network from scratch for 300K iterations. All experiments were conducted on a PC with an Nvidia GTX 970 GPU.

\subsection{Ablation Study}\label{section4_3}

In this section, we present ablation experiments on the Vid4 dataset to justify our design choices.

\begin{figure*}[bt]
	\centering
	\includegraphics[width=0.8\linewidth]{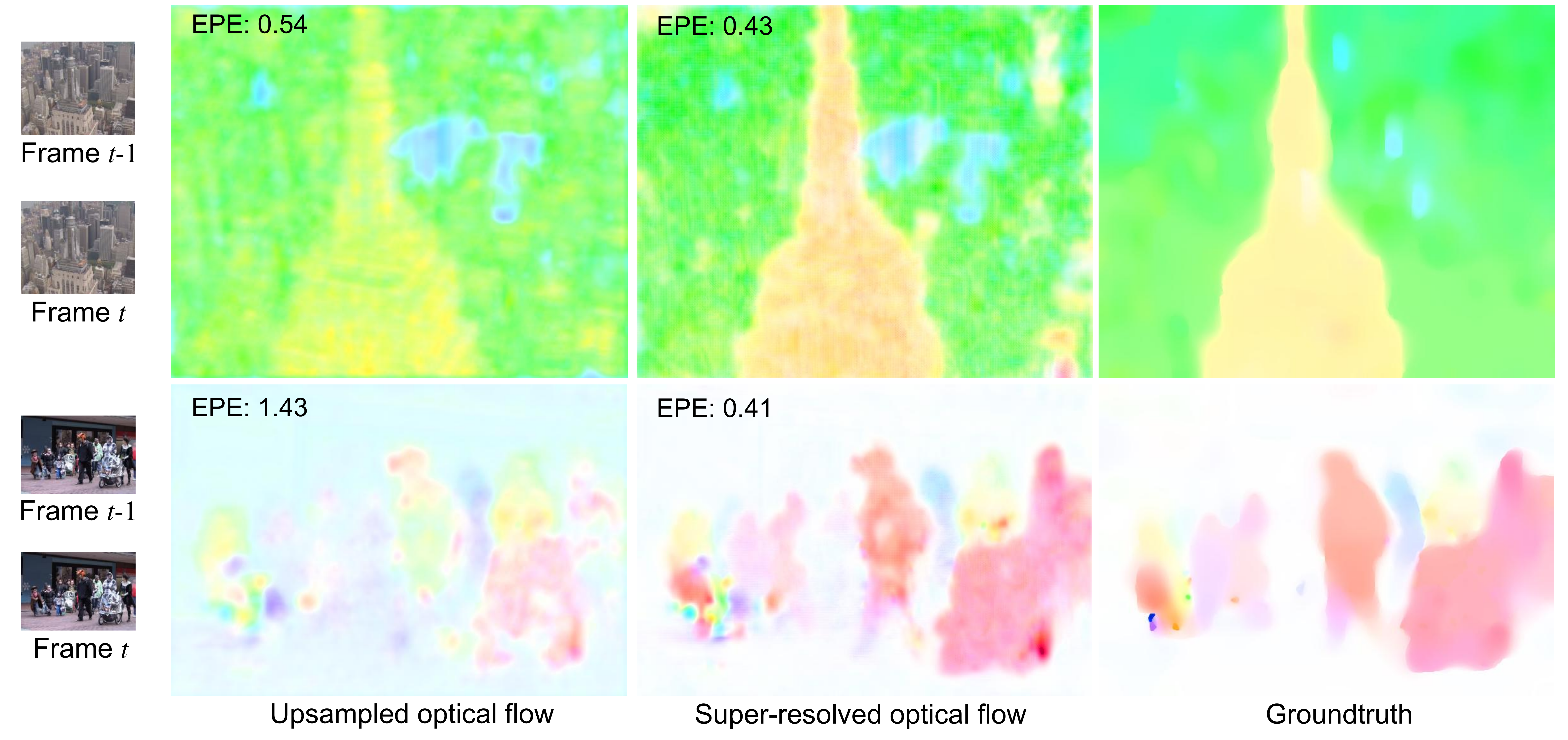}
	\caption{Visual comparison of optical flow estimation results achieved on \emph{City} and \emph{Walk} under $\times$4 configuration. The super-resolved optical flow recovers fine correspondences, which are consistent with the groundtruth.}
	\label{fig6}
\end{figure*}

\subsubsection{Network Variants} We proposed several variants of our SOF-VSR to perform ablation study. All the variants were re-trained for 300K iterations on the training data.

\textbf{SOF-VSR w/o OFRnet.} To handle complex motion patterns in video sequences, optical flow is used for motion compensation in our framework. To test the effectiveness of motion compensation for video SR, we removed the whole OFRnet and fed LR frames directly to our SRnet. Note that, replicated LR frames were used to match the dimension of the draft cube $C^{L}$.

\textbf{SOF-VSR w/o OFRnet$\mathcal{_\mathrm{\emph{level3}}}$.} The SR of optical flows provides accurate correspondences for video SR and improves the overall performance. To validate the effectiveness of HR optical flows, we removed the module at level 3 in our OFRnet. Specifically, the LR optical flows at level 2 were directly used for motion compensation and subsequent processing. To match the dimension of the draft cube, compensated LR frames were also replicated before feeding to SRnet.

\textbf{SOF-VSR w/o OFRnet$\mathcal{_\mathrm{\emph{level3}}}$ + upsampling.} Super-resolving the optical flow can also be simply achieved using interpolation-based methods. However, our OFRnet can recover more reliable optical flow details. To demonstrate this, we removed the module at level 3 in our OFRnet, and upsampled the LR optical flows at level 2 using bilinear interpolation. Then, we used the modules in our original framework for subsequent processing.

\subsubsection{Experimental Analyses}

To test the accuracy of individual output image, we used PSNR/SSIM as metrics. To further test the consistency performance, we used the temporal motion-based video integrity evaluation index (T-MOVIE) \cite{2010-MotionTunedSpatioTemporalQualityAssessmentofNaturalVideos-Seshadrinathan-335-350}. Besides, we used MOVIE \cite{2010-MotionTunedSpatioTemporalQualityAssessmentofNaturalVideos-Seshadrinathan-335-350} and video quality measure with variable frame delay (VQM-VFD) \cite{2011-VideoQualityModelforVariableFrameDelayVQMVFD-Wolf--} for overall evaluation. The MOVIE and VQM-VFD metrics are correlated with human perception and widely applied in video quality assessment. Evaluation results of our original framework and the 3 variants achieved on the Vid4 dataset are shown in Table~\ref{tab1}.

\textbf{Motion compensation.}
It can be observed from Table~\ref{tab1} that motion compensation plays a significant role in performance improvement. If OFRnet is removed, the PSNR/SSIM values are decreased from 26.01/0.771 to 25.80/0.760. Besides, the consistency performance is also degraded, with T-MOVIE value being increased to 20.08. That is because, it is difficult for SRnet to learn the non-linear mapping between LR and HR images under complex motion patterns.

\begin{table}[bt]
	\caption{Average EPE results achieved on the Vid4 dataset under $\times4$
		configuration. Best results are shown in boldface.}
	\label{tab2}
	\begin{center}
		\begin{tabular}{ccc}
			\hline
			& \tabincell{c}{Upsampled \\optical flow} & \tabincell{c}{Super-resolved \\ optical flow}\tabularnewline
			\hline 
			Calendar & 0.85 & \textbf{0.39} \tabularnewline
			City & 1.17 & \textbf{0.49} \tabularnewline
			Foliage & 1.18 & \textbf{0.36} \tabularnewline
			Walk & 1.25 & \textbf{0.55}\tabularnewline
			Average & 1.11 & \textbf{0.45} \tabularnewline
			\hline
		\end{tabular}
	\end{center}
\end{table}

\textbf{HR optical flow.}
If modules at levels 1 and 2 are introduced to generate LR optical flows for motion compensation, the PSNR/SSIM values are increased to 25.88/0.764. However, the performance is still inferior to our SOF-VSR method using HR optical flows. That is because, HR optical flows provide more accurate correspondences for performance improvement. If bilinear interpolation is used to upsample LR optical flows, no consistent improvement can be observed. That is because, upsampling operation cannot recover reliable correspondence details as the module at level 3. To demonstrate this, we further compared the super-resolved optical flow (output at level 3), upsampled optical flow (upsampling result of the output at level 2) to the groundtruth. Since no groundtruth optical flow is available for the Vid4 dataset, we used the method proposed by Hu \emph{et al.} \cite{2017-RobustInterpolationofCorrespondencesforLargeDisplacementOpticalFlow-Hu-4791-4799} to compute the groundtruth optical flow. We used the average end-point error (EPE) for quantitative comparison, and present the results in Table~\ref{tab2}. 

It can be seen from Table~\ref{tab2} that the super-resolved optical flow significantly outperforms the upsampled optical flow, with an average EPE being reduced from 1.11 to 0.45. It demonstrates that the module at level 3 effectively recovers the correspondence details. Figure ~\ref{fig6} further illustrates the qualitative comparison on \emph{City} and \emph{Walk}. In the upsampled optical flow, we can roughly distinguish the outlines of the building and the pedestrian. In contrast, more distinct edges can be observed in the super-resolved optical flow, with finer details being recovered. Although some checkboard artifacts generated by the sub-pixel layer can also be observed \cite{odena2016deconvolution}, the resulting HR optical flow provides highly accurate correspondences for the video SR task.

\subsection{Comparisons to the state-of-the-art}

We first compared our framework to IDNnet \cite{2018-FastandAccurateSingleImageSuperResolutionViaInformationDistillationNetwork-Hui--}
(the latest state-of-the-art single image SR method) and several video SR methods including VSRnet \cite{2016-VideoSuperResolutionwithConvolutionalNeuralNetworks-Kappeler-109-122},
VESCPN \cite{2017-RealTimeVideoSuperResolutionwithSpatioTemporalNetworksandMotionCompensation-Caballero-2848-2857},
DRVSR \cite{2017-DetailRevealingDeepVideoSuperResolution-Tao-4482-4490},
TDVSR \cite{2017-RobustVideoSuperResolutionwithLearnedTemporalDynamics-Liu--}
and FRVSR \cite{2018-FrameRecurrentVideoSuperResolution-Sajjadi--}
on the Vid4 dataset. Then, we conducted comparative experiments on the DAVIS-10 dataset.

\begin{table*}[bt]
	\caption{Comparison of accuracy and consistency performance achieved on the Vid4 dataset under $\times4$
		configuration. Note that, the first and last two frames are not used
		in our evaluation since VSRnet and TDVSR do not produce outputs for
		these frames. Results marked with * are directly copied from the corresponding papers. Best results are shown in boldface.}
	\label{tab3}
	\begin{center}
		\scriptsize
		\setlength{\tabcolsep}{0.6mm}{
			\begin{tabular}{ccccccccc}
				\hline 
				& \multicolumn{5}{c}{BI degradation model} & \multicolumn{3}{c}{BD degradation model}\tabularnewline 
				& \tabincell{c}{IDNnet \\ \cite{2018-FastandAccurateSingleImageSuperResolutionViaInformationDistillationNetwork-Hui--}}  & \tabincell{c}{VSRnet\\\cite{2016-VideoSuperResolutionwithConvolutionalNeuralNetworks-Kappeler-109-122}}  & \tabincell{c}{VESCPN\\\cite{2017-RealTimeVideoSuperResolutionwithSpatioTemporalNetworksandMotionCompensation-Caballero-2848-2857}}  & \tabincell{c}{TDVSR\\\cite{2017-RobustVideoSuperResolutionwithLearnedTemporalDynamics-Liu--}}  & SOF-VSR  & \tabincell{c}{DRVSR\\\cite{2017-DetailRevealingDeepVideoSuperResolution-Tao-4482-4490}}  & \tabincell{c}{FRVSR-3-64\\\cite{2018-FrameRecurrentVideoSuperResolution-Sajjadi--}}  & SOF-VSR-BD\tabularnewline
				\hline 
				PSNR($\uparrow$)  & 25.06  & 24.81  & 25.35*  & 25.49  & \textbf{26.01}  & 25.99  & 26.17*  & \textbf{26.19} \tabularnewline
				SSIM($\uparrow$)  & 0.715  & 0.702  & 0.756*  & 0.746  & \textbf{0.771}  & 0.773  & \textbf{0.798*}  & 0.785 \tabularnewline
				\tabincell{c}{T-MOVIE($\downarrow$)\\($\times10^{-3}$)} & 23.98 & 26.05 & - & 23.23 & \textbf{19.78} & 18.28 & - & \textbf{17.63} \tabularnewline
				\tabincell{c}{MOVIE($\downarrow$)\\($\times10^{-3}$)} & 5.99 & 6.01 & 5.82* & 4.92 &  \textbf{4.32} & \textbf{4.00} & - & \textbf{4.00}  \tabularnewline
				VQM-VFD($\downarrow$) & 0.268 & 0.273 & - & 0.238 & \textbf{0.227} & 0.217 & - & \textbf{0.215}\tabularnewline
				\hline
		\end{tabular}}
	\end{center}
\end{table*}

\begin{table*}[bt]
	\caption{Comparative results achieved on the DAVIS-10 dataset under $\times4$
		configuration. Best results are shown in boldface.}
	\label{tab4}
	\begin{center}
		\begin{tabular}{cccccc}
			\hline 
			& \multicolumn{3}{c}{BI degradation model} & \multicolumn{2}{c}{BD degradation model}\tabularnewline 
			& IDNnet\cite{2018-FastandAccurateSingleImageSuperResolutionViaInformationDistillationNetwork-Hui--}  & VSRnet\cite{2016-VideoSuperResolutionwithConvolutionalNeuralNetworks-Kappeler-109-122}  & SOF-VSR  & DRVSR\cite{2017-DetailRevealingDeepVideoSuperResolution-Tao-4482-4490} & SOF-VSR-BD\tabularnewline
			\hline 
			PSNR($\uparrow$) & 33.74 & 32.63 & \textbf{34.32} & 33.02 & \textbf{34.27} \tabularnewline
			SSIM($\uparrow$) & 0.915 & 0.897 & \textbf{0.925} & 0.911 & \textbf{0.925} \tabularnewline
			T-MOVIE($\times10^{-3}$)($\downarrow$) & 12.16 & 14.60 & \textbf{11.77} & 14.06 & \textbf{10.93} \tabularnewline
			MOVIE($\times10^{-3}$)($\downarrow$) & 2.19 & 2.85 & \textbf{1.96} & 3.15 & \textbf{1.90} \tabularnewline
			VQM-VFD($\downarrow$) & 0.146 & 0.163 & \textbf{0.119} & 0.142  & \textbf{0.127} \tabularnewline
			\hline
		\end{tabular}
	\end{center}
\end{table*}

For IDNnet and VSRnet, we used the codes provided by the authors to produce the results. For DRVSR and TDVSR, we used the output images provided by the authors. For VESCPN and FRVSR, the results reported in their papers \cite{2017-RealTimeVideoSuperResolutionwithSpatioTemporalNetworksandMotionCompensation-Caballero-2848-2857,2018-FrameRecurrentVideoSuperResolution-Sajjadi--} are used.
Here, we report the performance of FRVSR-3-64 since its network size is comparable to our SOF-VSR. 
Following \cite{2017-NTIRE2017ChallengeonSingleImageSuperResolution:MethodsandResults-Timofte-1110-1121}, we crop borders of $6+s$ for fair comparison. 

Note that, DRVSR and FRVSR are trained on a degradation model different from other networks. Specifically, the degradation model used in IDNnet, VSRnet, VESCPN and TDVSR is bicubic downsampling with Matlab $imresize$ function (denoted as BI). However, in DRVSR and FRVSR, the HR images are first blurred using Gaussian kernel and then downsampled by selecting every $s^{\textup{th}}$ pixel (denoted as BD). Consequently, we re-trained our framework on the BD degradation model (denoted as SOF-VSR-{BD}) for fair comparison to DRVSR and FRVSR.

Without optimization of the implementation, our SOF-VSR network takes about 250ms to
generate an HR image of size 720$\times$576 under $\times4$ configuration on
an Nvidia GTX 970 GPU.

\begin{figure}
	\centering
	\includegraphics[width=0.8\linewidth]{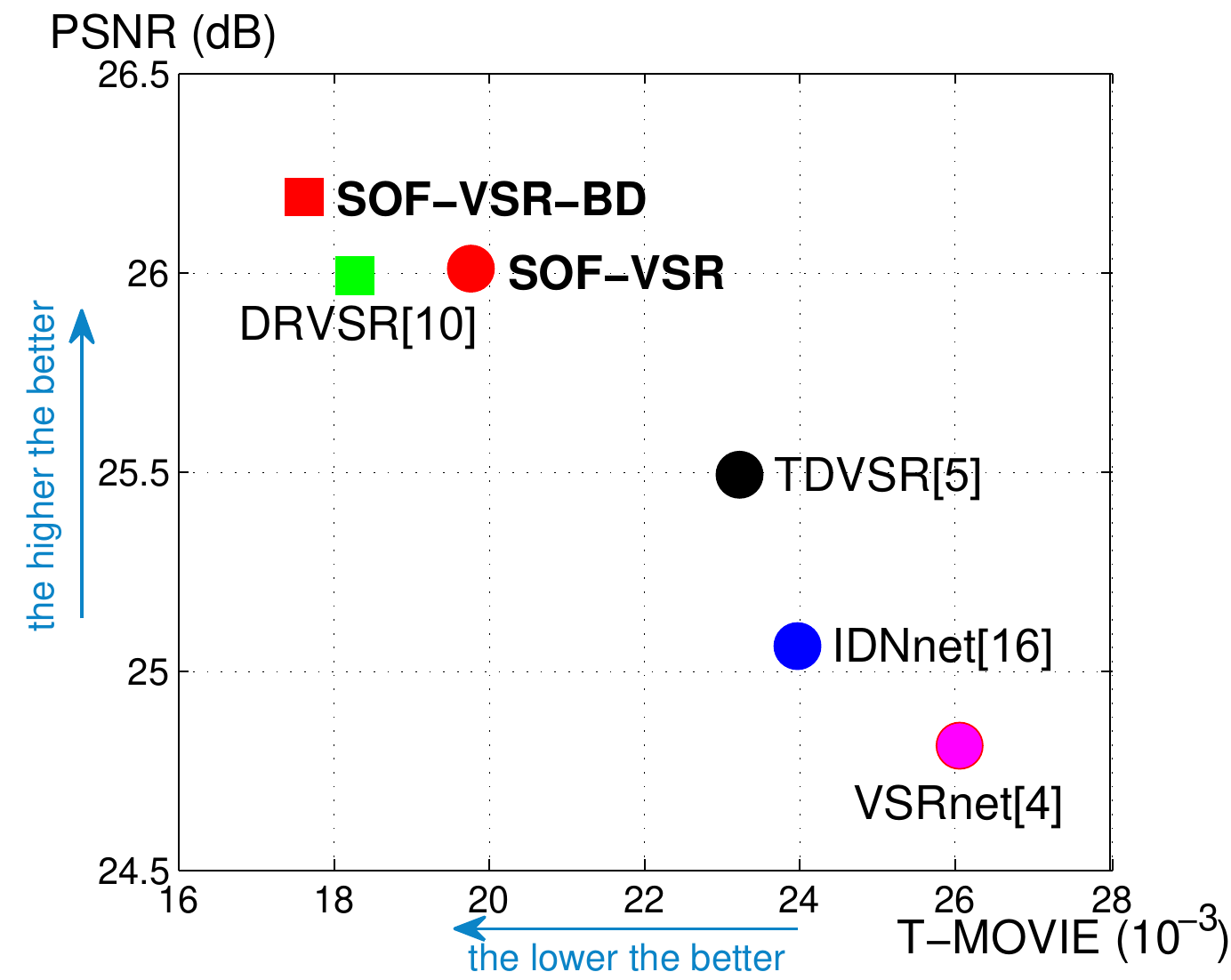}
	\caption{Consistency and accuracy performance achieved on the Vid4 dataset under $\times$4 configuration. Dots and squares represent performance for BI and BD degradation models, respectively. Our framework achieves the best performance in terms of both PSNR and T-MOVIE.}
	\label{fig7}
\end{figure}

\begin{figure*}
	\centering
	\includegraphics[width=0.9\linewidth]{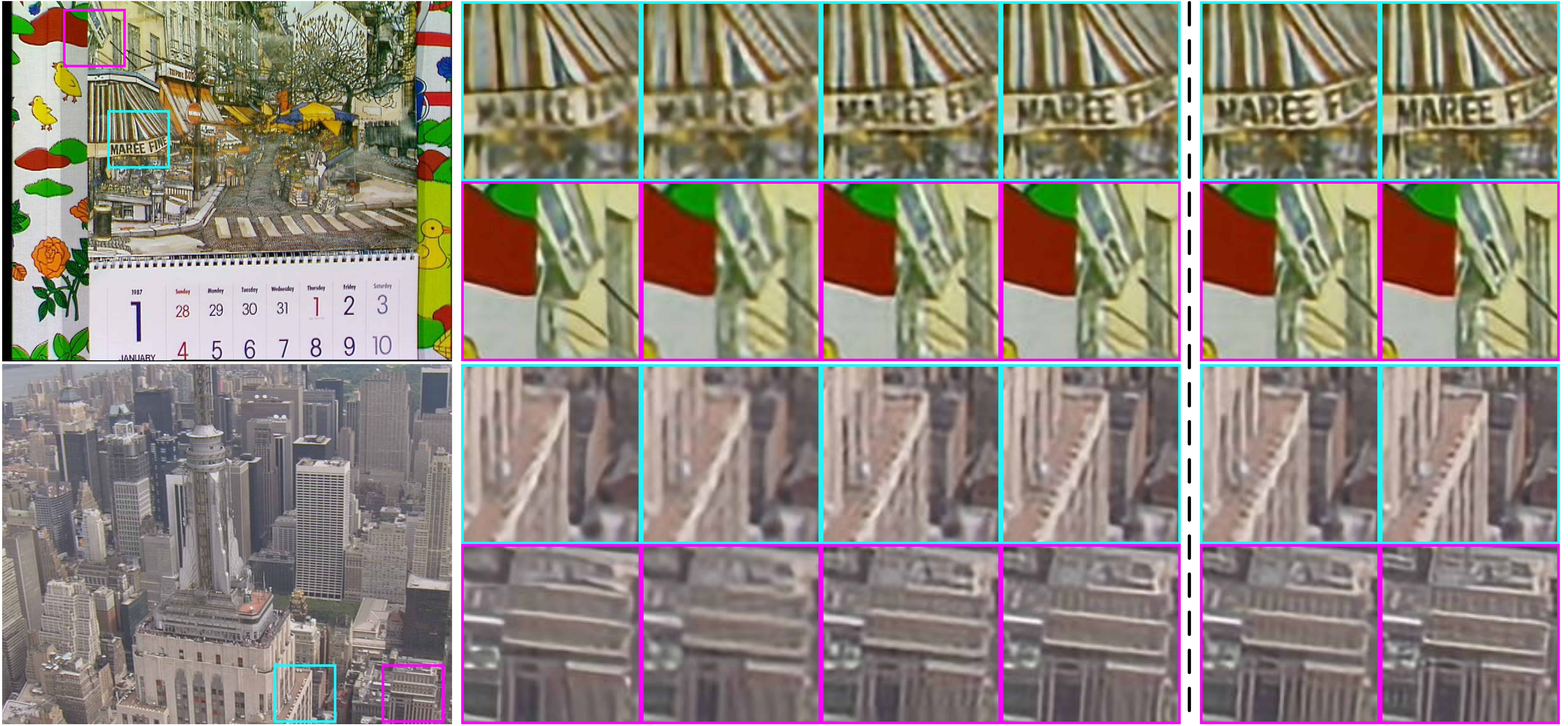}
	\caption{Visual comparison of $\times4$ SR results on \emph{Calendar} and \emph{City}. Zoom-in regions from left to right: IDNnet \cite{2018-FastandAccurateSingleImageSuperResolutionViaInformationDistillationNetwork-Hui--}, VSRnet \cite{2016-VideoSuperResolutionwithConvolutionalNeuralNetworks-Kappeler-109-122}, TDVSR \cite{2017-RobustVideoSuperResolutionwithLearnedTemporalDynamics-Liu--}, our SOF-VSR, DRVSR \cite{2017-DetailRevealingDeepVideoSuperResolution-Tao-4482-4490} and our SOF-VSR-BD. IDNnet, VSRnet, TDVSR and SOF-VSR are based on the BI degradation model, while DRVSR and SOF-VSR-BD are based on the BD degradation model.}
	\label{fig8}
\end{figure*}

\begin{figure*}
	\centering
	\includegraphics[width=0.9\linewidth]{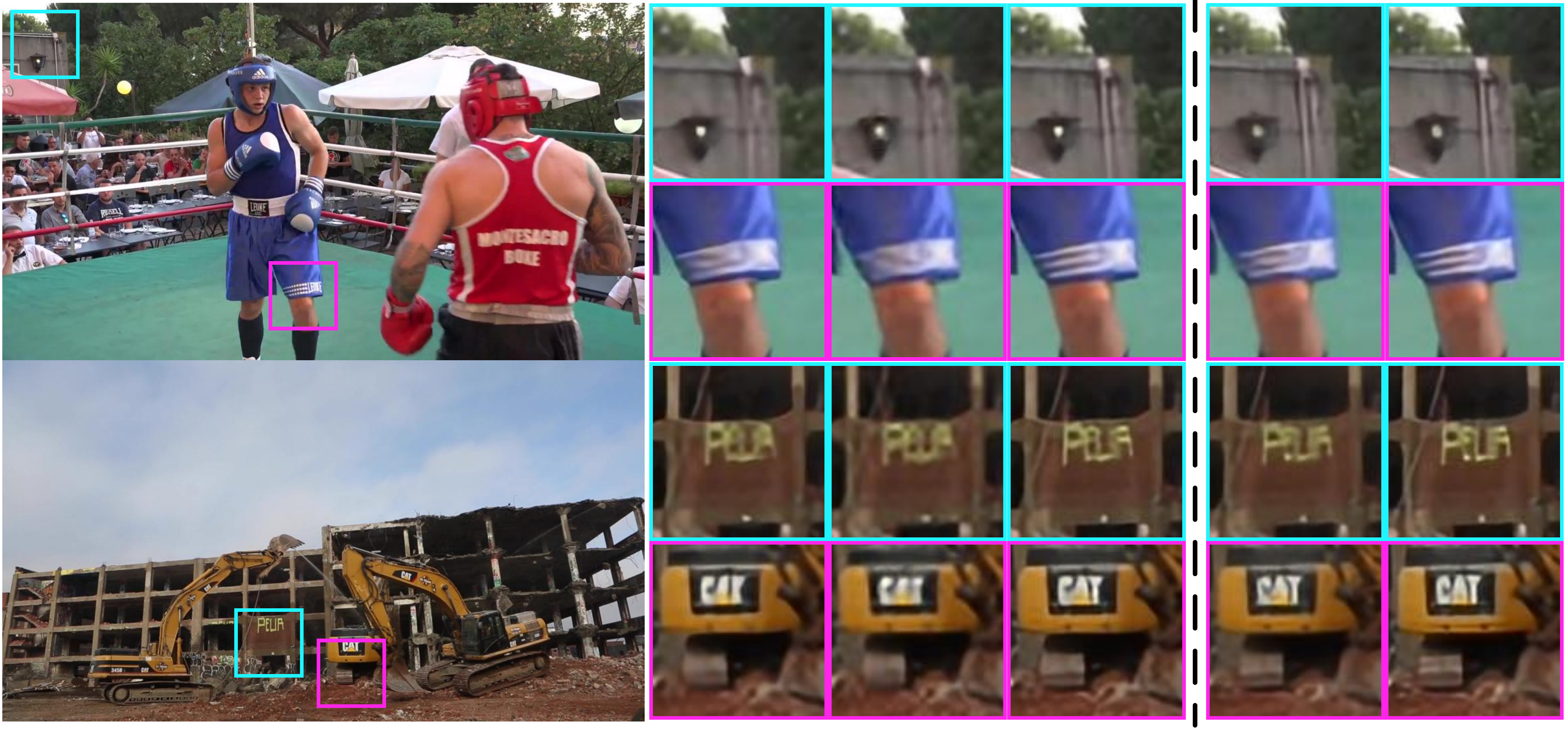}
	\caption{Visual comparison of $\times4$ SR results on \emph{Boxing} and \emph{Demolition}. Zoom-in regions from left to right: IDNnet \cite{2018-FastandAccurateSingleImageSuperResolutionViaInformationDistillationNetwork-Hui--}, VSRnet \cite{2016-VideoSuperResolutionwithConvolutionalNeuralNetworks-Kappeler-109-122}, our SOF-VSR, DRVSR \cite{2017-DetailRevealingDeepVideoSuperResolution-Tao-4482-4490} and our SOF-VSR-BD. IDNnet, VSRnet and SOF-VSR are based on the BI degradation model, while DRVSR and SOF-VSR-BD are based on the BD degradation model.}
	\label{fig10}
\end{figure*}

\subsubsection{Quantitative Evaluation} Quantitative results achieved on the Vid4 dataset and the DAVIS-10 dataset are shown in Tables~\ref{tab3} and ~\ref{tab4}.

\textbf{Evaluation on the Vid4 dataset.}
It can be observed from Table~\ref{tab3} that our SOF-VSR 
achieves the best performance for the BI degradation model in terms of all metrics. Specifically, the PSNR and SSIM values achieved by our framework are better than other methods by over 0.5 dB and 0.15 dB. That is because, more accurate correspondences can be provided by HR optical flows and therefore more reliable spatial details and temporal consistency can be well recovered. 

For the BD degradation model, although the FRVSR-3-64 method achieves higher SSIM, our SOF-VSR-{BD} method outperforms FRVSR-3-64 in terms of PSNR. Compared to the DRVSR method, PSNR, SSIM and T-MOVIE values achieved by our SOF-VSRBD method are improved by a notable margin, while a comparable performance is achieved in terms of MOVIE and VQM-VFD.

We further show the trade-off between consistency and accuracy in Fig.~\ref{fig7}. It can be seen that our SOF-VSR and SOF-VSR-BD methods achieve the highest PSNR values, while maintaining superior T-MOVIE performance.

\textbf{Evaluation on the DAIVIS-10 dataset.}
It is clear in Table~\ref{tab4} that our SOF-VSR and SOF-VSR-BD methods surpass the state-of-the-arts for both the BI and BD degradation models in terms of all metrics. Since the DAVIS-10 dataset comprises scenes with fast moving objects, complex motion patterns (especially large displacements) lead to deterioration of existing video SR methods. In contrast, more accurate correspondences are provided by HR optical flows in our framework. Therefore, complex motion patterns can be handled more robustly and better performance can be achieved.

\subsubsection{Qualitative Evaluation}

Figure ~\ref{fig8} illustrates the qualitative results on two scenarios of the Vid4 dataset. It can be observed from the zoom-in regions that our framework recovers finer and more reliable details, such as the word ``MAREE" and the stripes of the building. The qualitative comparison on the DAVIS-10 dataset (as shown in Fig.~\ref{fig10}) also demonstrates the superior visual quality achieved by our framework. The pattern on the shorts, the word ``PEUA" and the logo ``CAT"  are better recovered by our SOF-VSR and SOF-VSR-BD methods.

Figure ~\ref{fig1} further shows the temporal profiles achieved on \emph{Calendar} and \emph{City}. It can be observed that the word ``MAREE" can hardly be recognized by VSRnet in both image space and temporal profile. Although finer results are achieved by TDVSR, the building is still obviously distorted. In contrast, smooth and reliable patterns with fewer artifacts can be observed in temporal profiles of our results. In summary, our framework produces temporally more consistent results and better perceptual quality.

\section{Conclusions}

In this paper, we propose a deep end-to-end trainable video SR framework to super-resolve both images and optical flows. Our OFRnet first super-resolves the optical flows to provide accurate correspondences. Motion compensation is then performed based on HR optical flows and SRnet is used to infer the final results. Extensive experiments have demonstrated that our OFRnet can recover reliable correspondence details for the improvement of both accuracy and consistency performance. Comparison to existing video SR methods has shown that our framework achieves the state-of-the-art performance.

{\small
\bibliographystyle{ieee}
\bibliography{super-resolution,other-CV-fields,multi-image SR,datasets}
}

\end{document}